\documentclass[conference]{IEEEtran}
\bstctlcite{IEEEexample:BSTcontrol}
\ifCLASSINFOpdf

\else

\fi

\usepackage{graphicx}
\usepackage{subcaption}
\usepackage{float}

\usepackage{xcolor}

\usepackage{amsmath,amssymb,epsfig}

\usepackage{fancyhdr}
\usepackage{kantlipsum}
\fancypagestyle{plain}{
  \fancyhf{}
  \fancyhead[C]{Conference on \LaTeX}     
  \fancyfoot[L]{This is a notice}

}
\usepackage{eso-pic}
\begin{document}


\title{3D Object Localization Using 2D Estimates for Computer Vision Applications}
\author{\IEEEauthorblockN{Taha Hasan Masood Siddique\IEEEauthorrefmark{1},
Yawar Rehman\IEEEauthorrefmark{2},
Tahira Rafiq\IEEEauthorrefmark{4},
Malik Zohaib Nisar\IEEEauthorrefmark{3},\\
Muhammad Sohail Ibrahim\IEEEauthorrefmark{5}, and
Muhammad Usman\IEEEauthorrefmark{3}}
\IEEEauthorblockA{\\
\IEEEauthorrefmark{1}College of Information Science \& Electronic Engineering, Zhejiang University, Hangzhou-310027, Zhejiang, China.\\
Email: taha@zju.edu.cn\\
\IEEEauthorrefmark{2}Department of Electronic Engineering, NED University of Engineering and        Technology, Karachi-75270, Pakistan.\\
Email: yawar@neduet.edu.pk\\
\IEEEauthorrefmark{4}Department of Biomedical Engineering, Sungkyunkwan University, Suwon-16419, Republic of Korea.\\
Email: rtahira350@gmail.com \\
\IEEEauthorrefmark{5}College of Electrical Engineering, Zhejiang University, Hangzhou-310027, Zhejiang, China.\\
Email: msohail@zju.edu.cn\\
\IEEEauthorrefmark{3}Department of Computer Engineering, Chosun University, Gwangju-61452, Republic of Korea.\\
Email: \{zohaib, usman\}@chosun.kr\\
    }
}

\IEEEoverridecommandlockouts
\IEEEpubid{\makebox[\columnwidth]{978-1-6654-2413-4/21/\$31.00~\copyright2021 IEEE \hfill} \hspace{\columnsep}\makebox[\columnwidth]{ }}
 		
\maketitle

\IEEEpubidadjcol
\begin{abstract}
Computer vision based applications have received notable attention globally due to the interaction with the physical world. In this paper, a novel method for object localization based on camera calibration and pose estimation is discussed. The 3-dimensional (3D) coordinates are computed by taking multiple 2-dimensional (2D) images from different view that meets the requirement in camera calibration process. Several number of steps are involved in camera calibration including estimation of intrinsic and extrinsic parameters for the removal of lens distortion, estimation of object's size and camera location. Besides, a technique to estimate the 3D pose using 2D images is proposed and the results of camera parameters and localization are applied for the 3D reconstruction. The hardware implementation of the proposed approach is implemented on HP core i5 with the MATLAB support packages and experimental results are validated for both camera calibration and pose estimation.

\end{abstract}
	
	\IEEEpeerreviewmaketitle

\section{Introduction}
\label{intro}
In the era of technological advancement, computer vision has been playing a vital role in variety of fields including artificial intelligence, machine learning and image processing. These applications require spatial information of the objects to determine the physical axes and coordinates. Certain applications including interactive gaming devices and 3D printing, models the real-world objects, measures distance to provide virtual reality control \cite{wiley2018computer, 8954983}. For a highly precise and accurate experience, expensive 3D cameras with complicated configuration are mostly used \cite{giancola2018survey}. One off the shelf technique is 2D imaging, in which 2D images are captured from different angles and are transformed into 3D \cite{byun2017registration}. This method has been considered as a low cost alternative to acquire the said task. However, the performance is greatly dependent on the configuration and calibration of the camera \cite{usamentiaga2017highly}.

Camera calibration is indispensable in the computer vision applications, as most of the computer vision systems are highly affected by the precision of the calibration. The lenses of the camera are slightly skewed, due to which objects on the one side of the camera appear closer as compared to the other side causing distortion in captured image. Radial distortion in which the straight lines are appeared to be slightly curved is another kind of common distortion. Furthermore, in situations where the camera is not aligned to the image plane, leads to introduce tangential distortion. Calibration involves principal developments during 3D reconstructions including the restoration of camera's geometry \cite{sun2016camera}, extraction and analysis of 3D information and measurements with respect to the 3D world \cite{meng2017backtracking}. This technique is also employed towards estimation of the 3D location and the rotation of the camera relative to extrinsic and intrinsic parameters, which provide information about the 3D world coordinates and represents optical characteristics of camera respectively \cite{liu2017flexible}. A primary objective of camera calibration is the determination of the parameters of function, that explains the mapping from the position of a point in the 3D coordinates to the location of a point on the image plane \cite{fetic2012procedure}. 

In this research, we propose to use single camera calibration method where the camera projection matrix measures the 3D coordinates of a viewed points. The projection matrix is comprised of extrinsic and intrinsic parameters and converts the 3D object coordinates into 2D image coordinates. 
The rest of the paper is organized as follows: The related work has been discussed in section \ref{sec: RW}, followed by the methodology of the proposed work in section \ref{sec: Method}. Experimental results are presented in section \ref{sec3} and paper is finally  concluded in section \ref{sec:conclusion}.

\section{Related Work} \label{sec: RW}
The design objective of any 3D computer vision or machine vision is to extract image information from the camera and estimate the position, shape information and recognize 3D objects in the 3D world coordinates \cite{huang2019perspectivenet}. The geometric parameters, also known as camera parameters, empirically determines the relationship between the camera position and 3D world coordinates. Checkerboards are frequently used in camera calibration, allowing camera parameters to extract more precise information from images \cite{duda2018accurate}. A camera calibration method to reconstruct the 3D image of leaves using two 2D cameras has been presented in \cite{syahputra2017camera}. Camera calibration was essentially performed by utilizing the correlation methods on a particular region, followed by the extraction of intrinsic and extrinsic parameters by employing singular value decomposition. The resulting image however had rough patterns, therefore, a post-processing step of image smoothing was inevitable.  A 3D reconstruction method by using a flexible planar calibration plate has been proposed in \cite{gai2018novel}. Several images of the calibration plate at distinct angles are acquired  using two cameras. Centroid distance increment matrix is then applied to calculate the rotation and translation matrix which essentially links the frames of both cameras. To reconstruct the coordinates, space intersection method had been implemented the results of those are utilized to calculate the reconstruction error.  In \cite{urban2015improved}, a flexible technique that requires the camera to observe the object's pattern from a few angles only has been utilized. Furthermore, an improved method for calibrating an omni-directional imaging system is presented which helps to decrease the number of calibration steps. 
Currently, the camera calibration techniques are classified into two parts. (i) The target calibration method, which generally relates the known target information to the camera parameters and determines 2D to 3D relationship compatibility \cite{zhang2018single}. (ii) The self-calibration method, that does not require any targets and instead uses mathematical modeling to determine the camera parameters \cite{chang2018vanishing}. 

Mapping of camera position from 2D image to the 3D world coordinates is a common issue in image processing. When the 3D coordinate information is acquired by computer vision, the distortion coefficients and camera parameters e.g. the focal length is needed to be calculated beforehand. 
However, in \cite{rehman2019comparison}, the camera calibration is implied by developing the 3D point of scanning object, using the structure from motion technique without using any specific model of the camera. Similarly, checkerboard corners are intelligently used to estimate the pose of a camera \cite{shi2004new}. An approach for pose estimation with known camera parameters is presented in \cite{frahm2004pose}, which estimates the position of the multi-camera system using fixed orientations and translation of the camera. Another method for multi-camera calibration is presented in \cite{li2013multiple}, which resolves the checkerboard corners related problems using binary patterns between time and different cameras.

\section{Methodology}\label{sec: Method}
\subsection{Camera Model}
\label{sub1}
\begin{figure}[!ht]
	\begin{center}
		\centering
		\includegraphics[width=8cm]{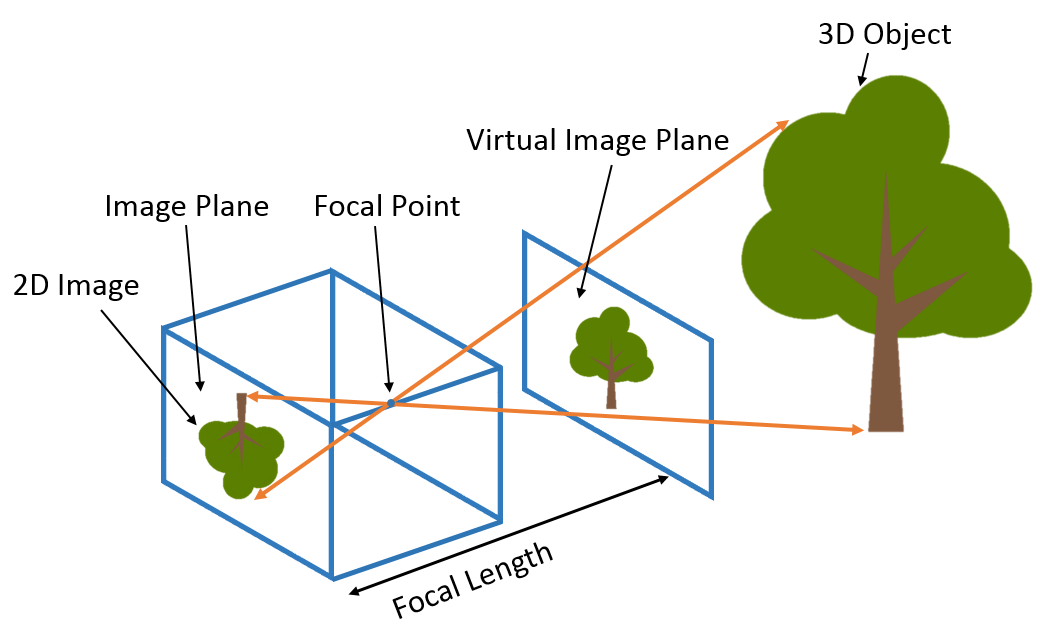}
	\end{center}
	\caption{Pin Hole Camera Model}
	
	\label{ref1}
\end{figure}

One of the crucial step in camera calibration and pose estimation is the selection of an appropriate modeling technique for the camera to obtain the intrinsic and extrinsic parameters. Internal parameters provide the geometry and optical characteristics of a camera including the focal length, image center and lens distortion where as the external parameters provide 3D orientation and position of a camera related to the world coordinates. These parameters are commonly used in pin hole camera which has no lens and contains single small aperture. When light rays enter into the aperture, an inverted image is formed on the opposite side of the camera which holds the erect reflection of the scene. This phenomenon has been shown in Figure \ref{ref1}.

The camera parameters consist of $4 \times 3$ matrix termed as the camera matrix. The camera matrix transforms the 3D coordinates into the 2D image points by using extrinsic and intrinsic parameters. The extrinsic parameters contains the location of the camera whereas intrinsic parameters holds focal length of the camera in the 3D world units as shown in equations below.

\begin{eqnarray}\label{eq1}
w
\begin{bmatrix}
    a & b  & 1  
  \end{bmatrix}
  =
  \begin{bmatrix}
    A & B & C &1  
  \end{bmatrix}P
\end{eqnarray}
where $w$ is a scaling factor, [$a$ $b$ $1$] are 2D image points and [$A$ $B$ $C$ $1$] are 3D unit matrix.

\begin{eqnarray}
P =
\begin{bmatrix}
    R  \\    t
  \end{bmatrix} K
\end{eqnarray}

where $P$ represents the camera matrix, $R$ and $t$ are the rotational and translation elements of the extrinsic matrix respectively, while $K$ represents the intrinsic matrix.
 
\begin{figure}[!ht]
	\begin{center}
		\centering
		\includegraphics[width=8cm]{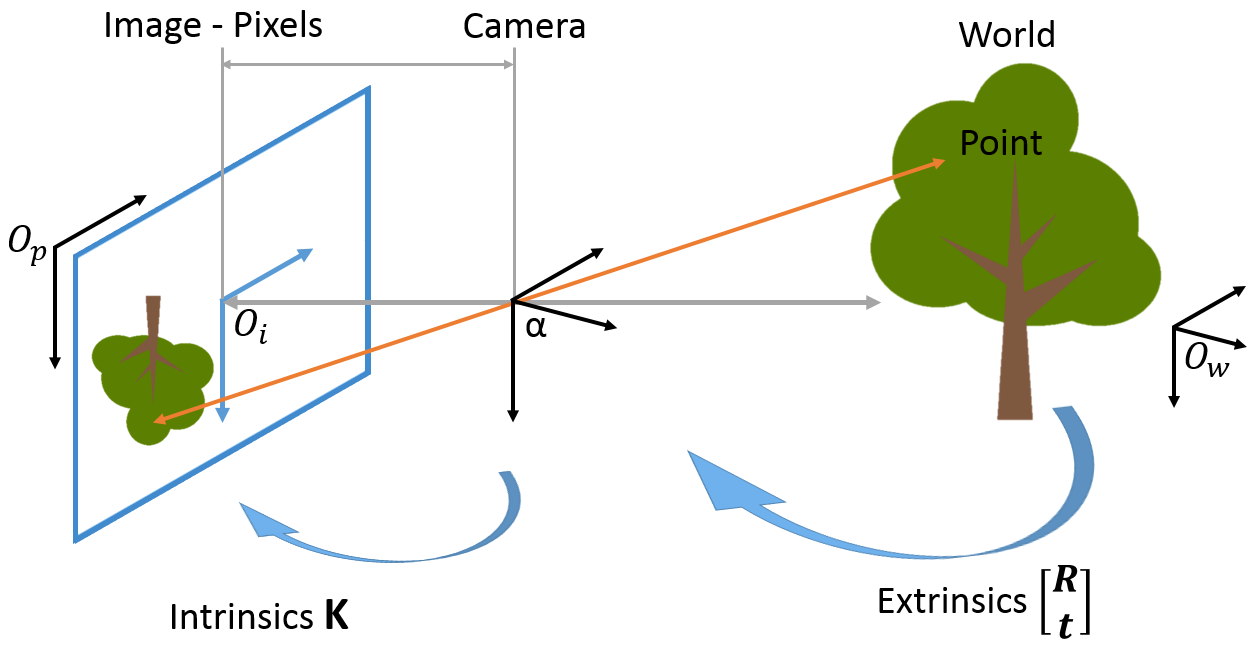}
	\end{center}
	\caption{Mapping of Camera Coordinates}
	
	\label{fig2}
\end{figure}

Extrinsic parameters transforms the 3D coordinates into camera coordinates and then intrinsic parameters converts the camera coordinates into the image plane as shown in Figure \ref{fig2}.

\subsection{Calibration Method}
\label{sub2}
\begin{figure}[!ht]
	\begin{center}
		\centering
		\includegraphics[width=8cm]{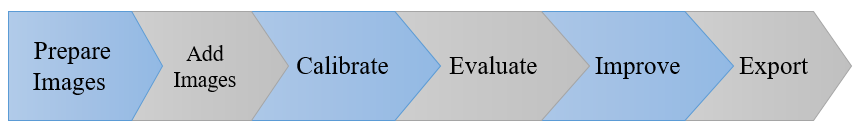}
	\end{center}
	\caption{Camera Calibration Workflow}
	\label{cap}
\end{figure}

We present the calibration technique to calibrate an off the shelf Logitech HD c270 USB webcam. The camera calibration is carried out on HP core i5 computer with the MATLAB USB support package. Camera calibrator application is used to determine the camera's intrinsic, extrinsic and lens distortion parameters. These parameters can be used for various computer vision application including removal of lens distortion from an image, measuring the object's size or reconstructing 3D scenes from multiple position. Once the camera is connected to the computer, a set of checkerboard test images must be added. The calibration algorithm detects the corners of checkerboard to ensure that test images to meet the calibrator requirement. When the test image satisfies the calibrator requirement, the calibration accuracy can be evaluated and improved by analyzing the re-projections errors, camera's extrinsic parameters and viewing the un-distorted image. After successful calibration, the export camera parameters can be used for many computer vision task. Several methods including Scale Invariant Feature Transform (SIFT) \cite{lowe2004distinctive} and Speeded Up Robust Transform (SURF) \cite{bay2006surf} have been proposed that use the checkerboard corner detection for the extraction of checkerboard crossings square corners of checkerboard. The camera calibration workflow is shown in Figure \ref{cap}.

\subsection{Camera Calibration and Pose Estimation}
\label{sub3}
\begin{figure}[!ht]
	\begin{center}
		\centering
		\includegraphics[scale=0.35]{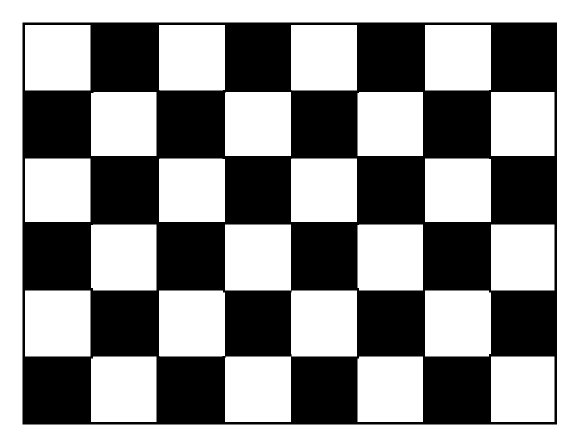}
	\end{center}
	\caption{Calibration Template}
	
	\label{fig3}
\end{figure}

An image of $10 \times 7$ checkerboard is used as a reference to obtain features of interest of the target image. The checkerboard pattern  must contain same number of black squares along one side and same number of white squares on the opposite side. This standard determines the orientation of checkerboard pattern in calibration. A 23mm checkerboard square size is used for calibration, a template of which is shown in Figure \ref{fig3}.

\begin{figure}[!ht]
	\begin{center}
		\centering
		\includegraphics[width=8cm]{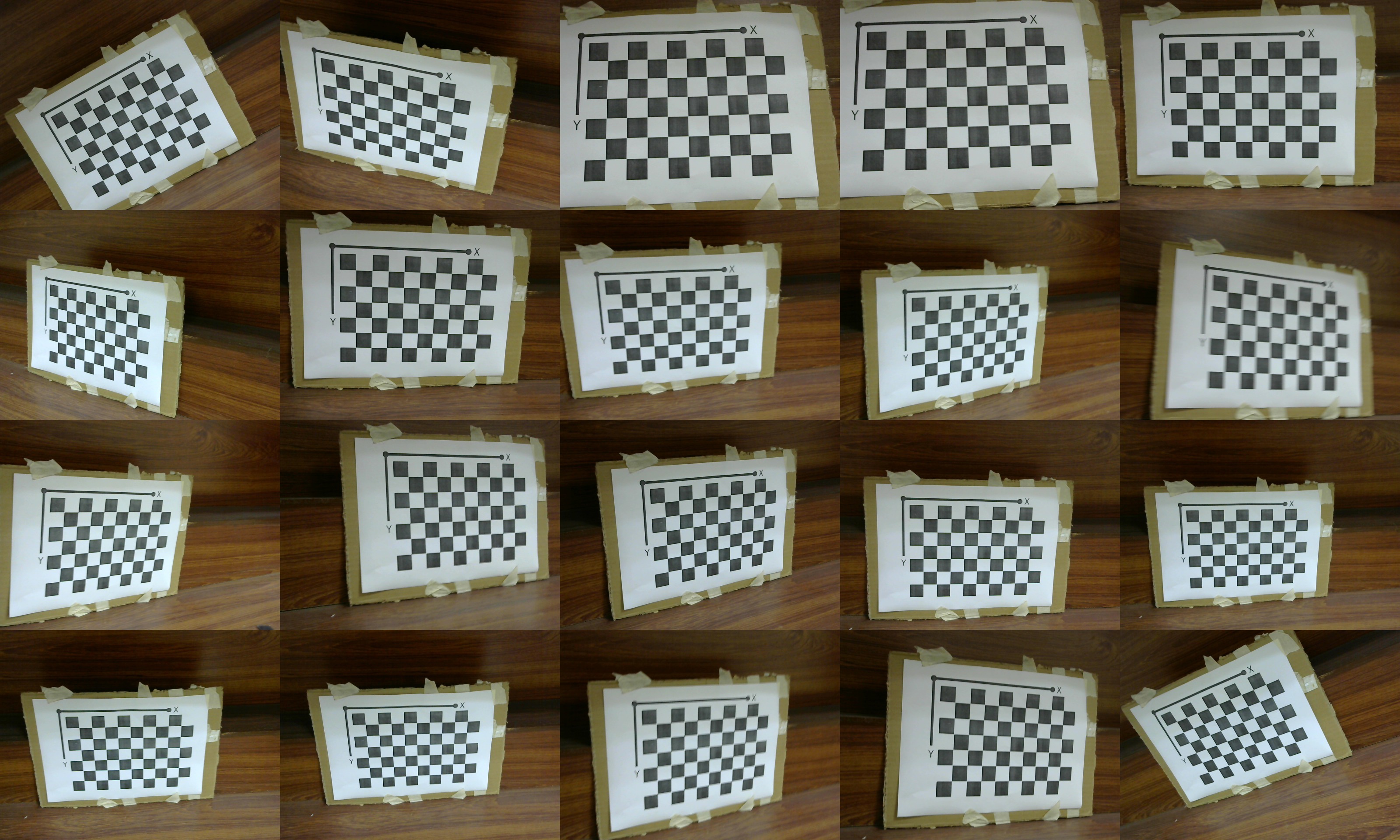}
	\end{center}
	\caption{Input Images in Calibration Process}
	\label{fig4}
\end{figure}

\begin{figure}[!ht]
	\begin{center}
		\centering
		\includegraphics[width=8cm]{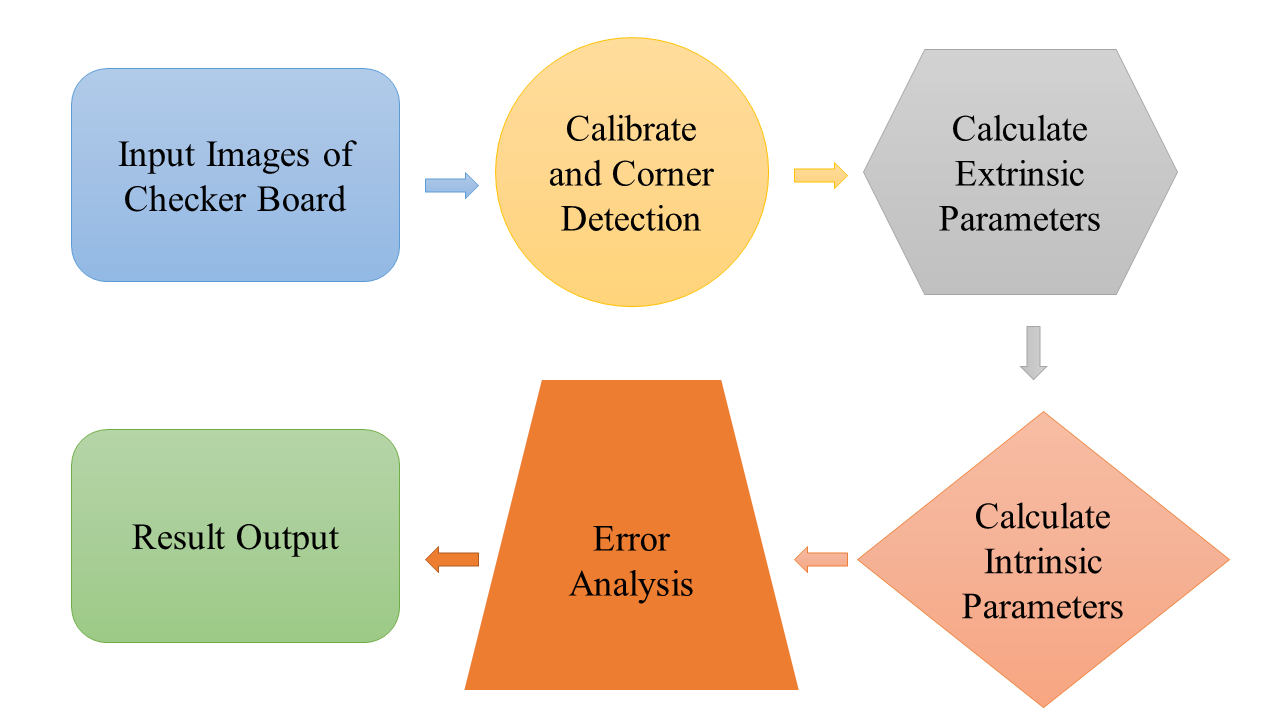}
	\end{center}
	\caption{Calibration Process Flow}
	\label{fig5}
\end{figure}

\begin{figure}[!ht]
	\begin{center}
		\centering
		\includegraphics[width=8cm]{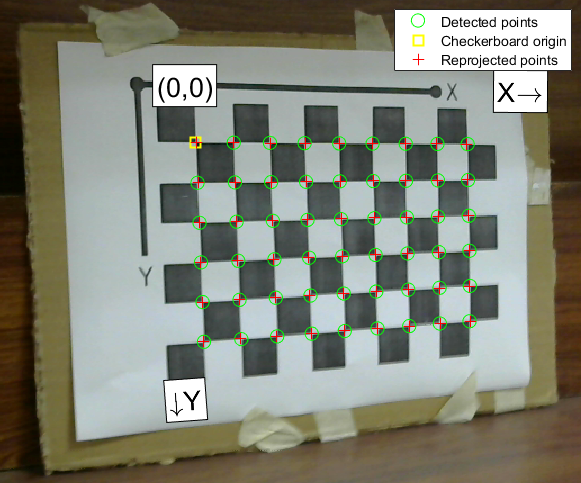}
	\end{center}
	\caption{Detected Corner Points of Checkerboard }
	\label{fig6}
\end{figure}

\begin{figure}[!ht]
	\begin{center}
		\centering
		\includegraphics[scale=0.35]{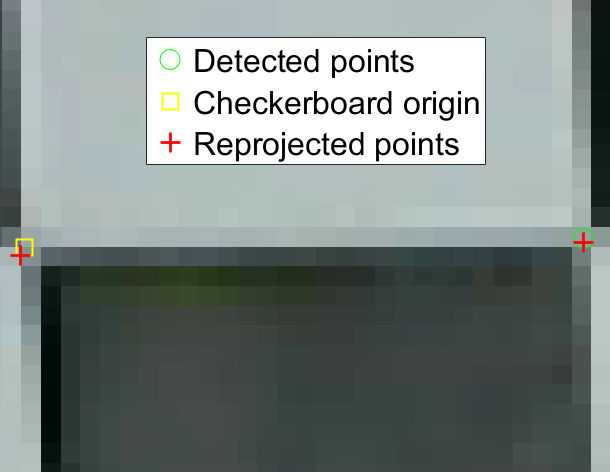}
	\end{center}
	\caption{Projection Error of Detecting Square Box}
	\label{fig7}
\end{figure}

Initially,  the input images are processed through MATLAB camera calibrator application. For better results and the evaluation of the corners, 20 images of calibration pattern are used which are shown in Figure \ref{fig4}. The default configuration calculates both camera parameters i.e intrinsic and extrinsic parameters depending upon the type of camera and then it computes the projection error. Once the calibration process is completed, the camera pose can be estimated with respect to the 3D object location as depicted in the calibration process flow in Figure \ref{fig5}. The detected corner points of checkerboard and the projection error of detecting square box of checkerboard are shown in Figure \ref{fig6} and Figure \ref{fig7} respectively. 

\section{Experimental Results}\label{sec3}


\subsection{Simulated Experiment}
\label{sec3a}
\begin{figure}[!ht]
 \begin{subfigure}{.20\textwidth}
\centering
\includegraphics[width=1.1\linewidth]{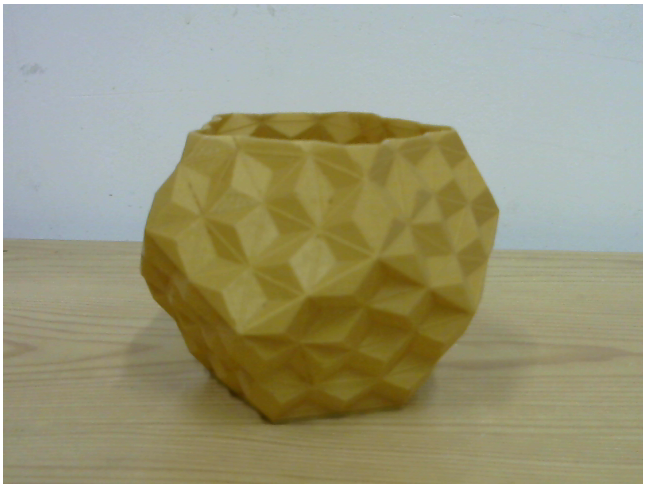}
\caption{Input Image}
\label{fig8a}
\end{subfigure}
\qquad
\begin{subfigure}{.20\textwidth}
\centering
\includegraphics[width=1.1\linewidth]{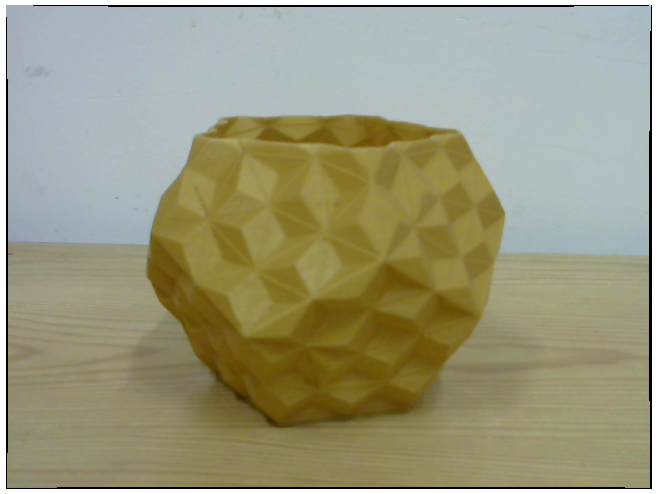}
\caption{Undistorted Image}
\label{fig8b}
\end{subfigure}
\caption{Removal of Distortion}
\end{figure}

\begin{figure}[!ht]
	\begin{center}
		\centering
		\includegraphics[width=8cm]{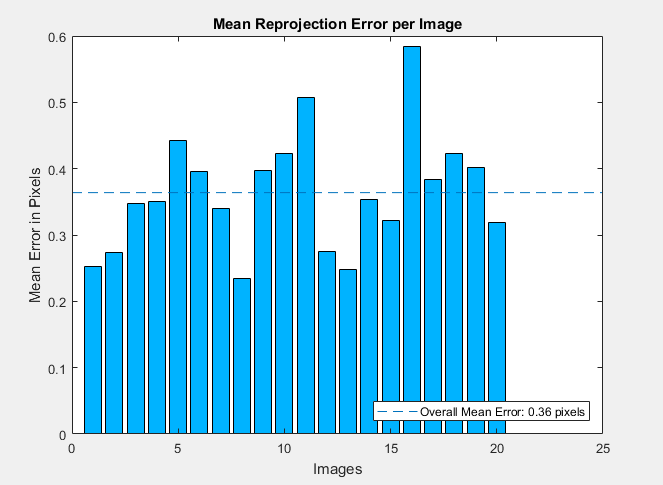}
	\end{center}
	\caption{Re-projection Errors}
	\label{fig11}
\end{figure}

From the simulation results, camera parameters  including focal length, principal point, radial distortion, mean projection error and intrinsic parameters matrix are calculated. The focal length, principal point and radial distortion are stored in a $2 \times1$ vector. The intrinsic parameters are stored in the $3\times3$ matrix along with mean projection error is calculated. These parameters are enlisted in Table \ref{TS:CORR}.
The calibration accuracy is examined by an un-distorted image,  camera extrinsic and re-projection errors. Figure \ref{fig8a} shows the input image which is taken from USB webcam. After removing lens distortion, the un-distorted image is obtained as shown in Figure \ref{fig8b}. The re-projection errors are the distances between the detected corner points and re-projected points of checkerboard in pixels. Generally, the mean re-projection error of less than one pixel is acceptable \cite{websitecite2}. The mean re-projection error per image in pixel and overall mean error of selected images is shown in Figure \ref{fig11}.

\begin{table}[!ht]
	\begin{center}
		\caption{Camera Parameters}
		\label{TS:CORR}
		\begin{tabular}{|c|c|}
			\hline Image size (pixels) & [480 640]\\
			\hline Focal length (pixels) & [  839.3458 $\pm$ 3.6694,   839.5573 $\pm$ 3.7166  ] \\
			\hline  Principal point (pixels) & [ 332.3661 $\pm$ 1.4489,      259.5099 $\pm$ 1.5829 ] \\
			\hline  Radial distortion & [ 0.0101 $\pm$ 0.0167   -0.1883 $\pm$ 0.1895 ]\\ 
			\hline Mean projection error & 0.363945706962709\\
			\hline Intrinsic parameters & 
			
			$\begin{bmatrix}
            839.345758 & 0.000000 & 0.000000 \\
            0.000000 & 839.557331  & 0.000000  \\
            332.366095 & 259.509924 & 1.000000
  \end{bmatrix}$ \\
  
  \hline
		\end{tabular}
	\end{center}
\end{table}

\begin{figure}[!ht]
	\begin{center}
		\centering
		\includegraphics[width=8cm]{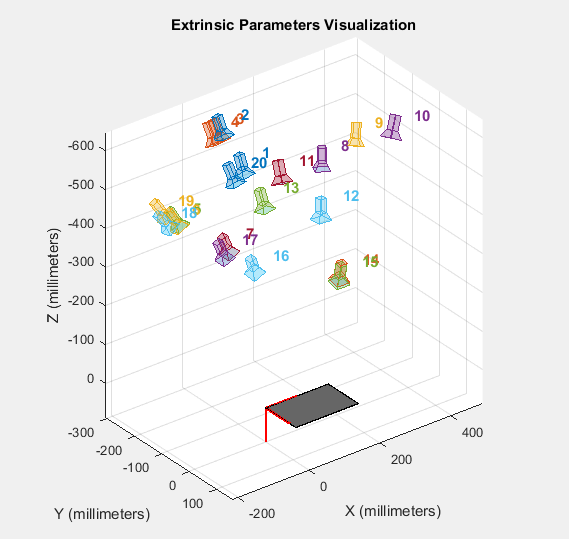}
	\end{center}
	\caption{Pattern View Visualization}
	\label{fig12}
\end{figure}

\begin{figure}[!ht]
	\begin{center}
		\centering
		\includegraphics[width=8cm]{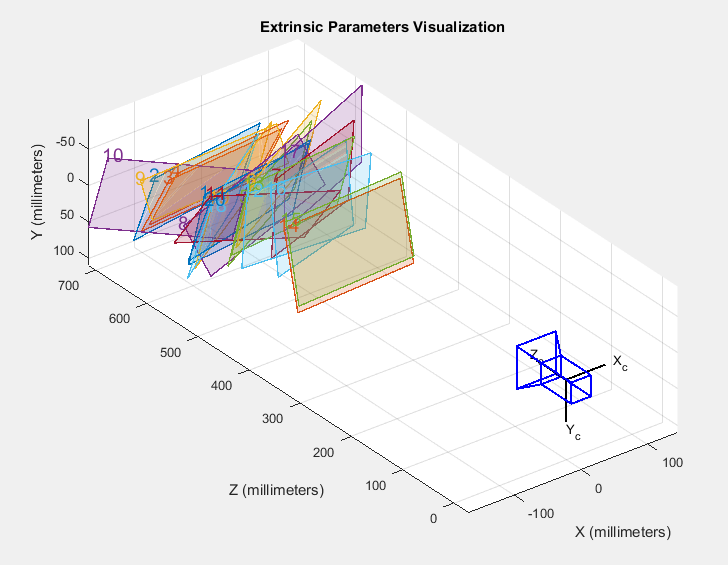}
	\end{center}
	\caption{Camera View Visualization}
	\label{fig131}
\end{figure}

The 3D extrinsic parameters plot provides a pattern and camera view visualization. The pattern view visualization is useful if the pattern is placed at fixed position. Similarly, the camera view visualization is useful when the images are captured by placing a camera at fixed position. The 3D extrinsic parameter visualization of pattern view visualization is shown in Figure \ref{fig12} and camera view visualization is shown in Figure \ref{fig131}.

\subsection{3D Reconstruction using Structure from Motion (SFM)}
\label{sec3b}
\begin{figure}[!ht]
	\begin{center}
		\centering
		\includegraphics[width=7cm]{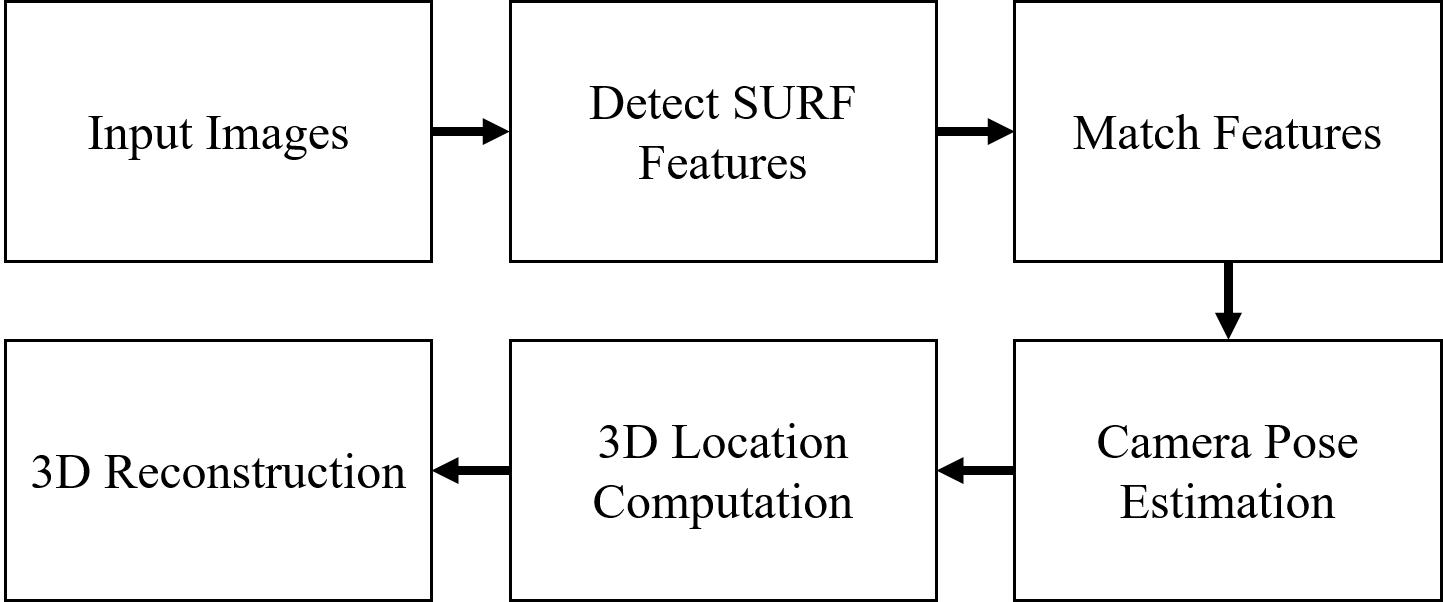}
	\end{center}
	\caption{Process Flow of Structure from Motion}
	
	\label{fig13}
\end{figure}

\begin{figure}[ht]
 \begin{subfigure}{.23\textwidth}
\centering
\includegraphics[width=\textwidth]{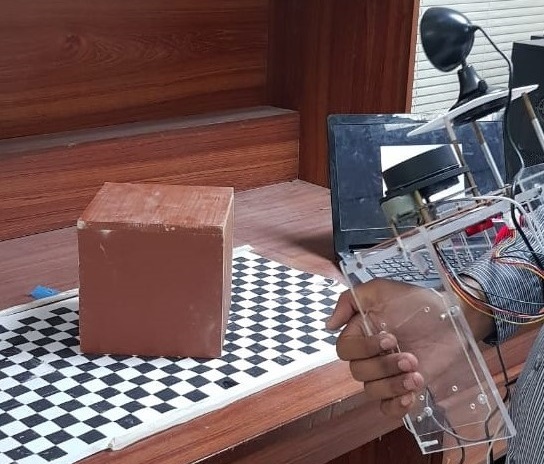}
\caption{Experimental Setup}
\label{fig3a}
\end{subfigure}
\quad
\begin{subfigure}{.23\textwidth}
\centering
\includegraphics[width=\textwidth]{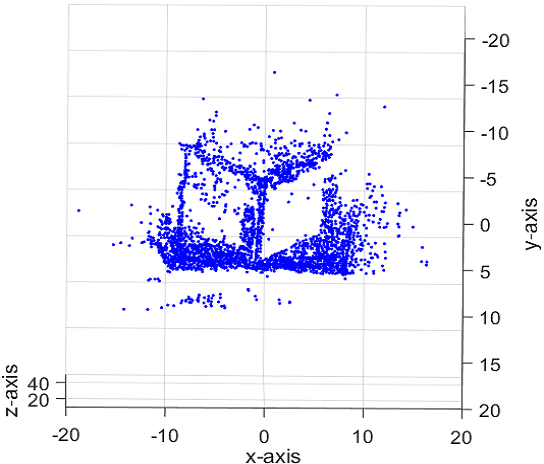}
\caption{3D Point Cloud}
\label{fig3b}
\end{subfigure}
\caption{Image to 3D Point Cloud Generation}
\end{figure}

Structure from Motion (SFM) is the method of estimating the 3D scene of an object by using multiple 2D images captured by a camera \cite{schonberger2016structure}. It is used in many applications such as 3D scanning \cite{daneshmand20183d}, augmented reality \cite{bae2016fast}, robot mapping \cite{saputra2018visual}, and autonomous driving \cite{fan2019key}. This algorithm takes input of multiple images and produce a series of output images called point cloud. By using SFM, the pose of the calibrated camera is estimated from a set of 2D images, and the information is applied to reconstruct the 3D point cloud of an unknown object. This algorithm consists of two parts, (i) camera pose estimation and (ii) 3D dense reconstruction. In the first part, the algorithm uses SURF technique to detect the interest points and features in the set of 2D images which compares the pairwise match to estimate the current view of camera pose related to the previous view. In the second part, SFM need interest points to be matched in multiple images commonly called track points. Once the relative position of camera from multiple views is known, then the track points serve as input to multiple images by using triangulation to compute 3D points. Using information from multiple views to get more precise estimation of the 3D points is called bundle adjustment. These 3D points are used in 3D reconstruction to refine the 3D world points and camera poses. The experiment is carried out on the cubic box to generate the 3D point cloud of the known object. Figure \ref{fig13} shows the process of structure from motion, experimental setup can be seen in Figure \ref{fig3a} and the 3D point cloud of cubic box is illustrated in Figure \ref{fig3b}.

\section{Conclusion}\label{sec:conclusion}
Object localization has several commercial and industrial application. In this work, a technique for  object localization  based  on camera calibration and pose estimation is discussed. The 3-dimensional (3D) coordinates are computed by taking multiple 2-dimensional (2D) images of the object from different views that meets the requirement of camera calibration and then the location of the camera in the 3D coordinate system is estimated by utilizing the information extracted from the 2D images of the checkerboard taken from several angles. The results of the camera parameters and localization are applied to the 3D reconstruction of the cubic box by using Struture from Motion (SFM) algorithm. The proposed technique achieves a mean projection error of less than one pixel re-projection error implying the successful calibration of Logitech HD c270 camera.

	\bibliographystyle{IEEEtran}
	\bibliography{Reference}

\end{document}